\documentclass[twoside,11pt]{article}

\usepackage[abbrvbib,preprint]{jmlr2e}

\usepackage{algorithm}
\usepackage{algpseudocode}
\usepackage{amsmath,amsfonts}
\usepackage{booktabs}

\newcommand{\T}{\hspace{3mm}}

\DeclareMathOperator*{\argmax}{argmax}

\ShortHeadings{Reward Shaping for Human Learning}{Rucker, Watson, Gerber and Barnes}
\firstpageno{1}

\begin{document}

\title{Reward Shaping For Human Learning\\ via Inverse Reinforcement Learning}

\author{\name Mark Rucker \email mr2an@virginia.edu \\
       \addr Department of Systems Engineering\\
       University of Virginia\\
       Charlottesville, VA 22904, USA
       \AND
       \name Layne T. Watson \email  ltwatson@computer.org \\
       \addr Departments of Computer Science, Mathematics, and
      Aerospace and Ocean Engineering\\
       Virginia Polytechnic Institute and State University\\
       Blacksburg, VA 24061, USA
       \AND
       \name Matthew S. Gerber \email msg8u@virginia.edu \\
       \addr Department of Systems and Information Engineering \\
       University of Virginia\\
       Charlottesville, VA 22904, USA
       \AND
       \name Laura E. Barnes \email lb3dp@virginia.edu \\
       \addr Department of Systems and Information Engineering \\
       University of Virginia\\
       Charlottesville, VA 22904, USA}

\editor{My editor}

\maketitle



\begin{abstract}%
Humans are spectacular reinforcement learners, constantly learning from and adjusting to experience and feedback. Unfortunately, this doesn't necessarily mean humans are fast learners. When tasks are challenging, learning can become unacceptably slow. Fortunately, humans do not have to learn tabula rasa, and learning speed can be greatly increased with learning aids. In this work we validate a new type of learning aid -- reward shaping for humans via inverse reinforcement learning (IRL). The goal of this aid is to increase the speed with which humans can learn good policies for specific tasks. Furthermore this approach compliments alternative machine learning techniques such as safety features that try to prevent individuals from making poor decisions. To achieve our results we first extend a well known IRL algorithm via kernel methods. Afterwards we conduct two human subjects experiments using an online game where players have limited time to learn a good policy. We show with statistical significance that players who receive our learning aid are able to approach desired policies more quickly than the control group.

\end{abstract}


\section{Introduction}
Spare the rod, spoil the child. The impact of environmental feedback on human behavior has long been understood at an intuitive level. However, it wasn't until the pioneering work of Thorndike \cite{thorndike1927law} and Pavlov \cite{clark2004classical} in the late nineteenth century that such intuition began to be formulated as a scientific system of knowledge. Fast forward to the present and much of the state of the art in machine learning draws inspiration from Thorndike's and Pavlov's theories in research area known as \textit{reinforcement learning} (RL).

In the classic research of Thorndike and Pavlov environmental feedback signals were known. What was not known is how these signals would influence animal behavior. Similarly, in RL research, environmental feedback signals (called reward functions) are also known. What is not known is the optimal behavior (called an optimal policy) for pursuing the reward function. For this reason traditional RL research focuses on methods to learn optimal policies from reward functions. More recently, some researchers have begun to turn the RL problem around with \textit{inverse reinforcement learning} (IRL). In the IRL problem a behavior policy is observed and the reward being pursued by the observed behavior needs to be learned.

With the IRL formulation in mind the authors wondered if IRL could be used to recover a shaped reward from a human expert for faster human task learning. Shaped reward functions increase learning speed by providing positive feedback for every action that moves a learning agent closer to an optimal policy \cite{peterson2004day}. This increased feedback is meant to guide learners towards optimal policies more quickly. Many algorithms have been proposed for IRL but in order to achieve reward shaping we selected a max-margin approach \cite{abbeel2004apprenticeship}. In max-margin IRL the goal is to recover a reward that makes the expert behavior appear maximally different from all other behavior. This means there is a steep ascent in reward received as one gets closer and closer to the expert behavior that generated the reward.

\begin{figure}
    \centering
    \includegraphics[width=1\textwidth]{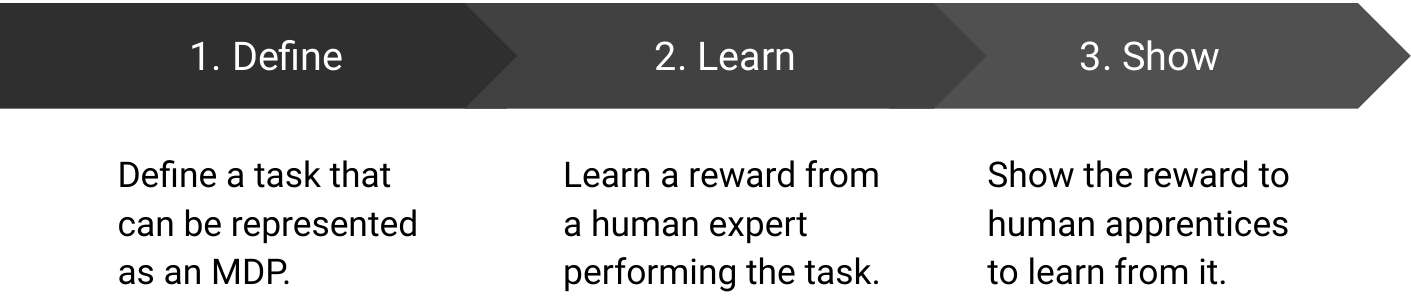}
    \caption{The proposed three step process used in this paper. \label{fig:proc}}
\end{figure}

In totality we propose a three step process to produce a learning aid for humans learning to perform sequential decision problem tasks, which we outline in Figure~\ref{fig:proc}. First, define the task that human reward shaping will be applied to. Second, learn reward functions from human's that are experts at the defined task using a max-margin IRL algorithm. Finally, show the learned reward functions to future human learners in order to increase their learning rate.


To test the proposed process two human subject experiments were conducted using an online game as our defined task. We then recruited a cohort to play the game and identified two experts in this cohort for reward learning. Finally, we used the learned rewards to modify the points future players received while playing the game. Players who received our reward shaping intervention demonstrated higher levels of performance when compared to a control group given equal amounts of time to play the game.

In what follows, Section~\ref{sec:related} gives a brief review of related work. Section~\ref{sec:prelim} provides a very cursory review of preliminary mathematical variables and equations. Section~\ref{sec:algorithm} develops the ``large-margin" IRL algorithm used in the experiments. Section~\ref{sec:experiment} describes the human subject experiments and their results. And Section~\ref{sec:conclusion} gives a brief summary of the contributions of this paper along with future research directions.

\section{Related Work\label{sec:related}}



Existing research applying IRL to human behavior can be separated into three domains: (1) human-behavior prediction, (2) human-cognition explanations and (3) human-machine collaboration. Despite each branch's distinct purpose they all share a similar motivation for using IRL. That is, that IRL has shown itself to be a strong candidate for modeling human decision making in a generalizable and computationally useful manner. 

The first domain, human-behavior prediction, seeks to build models of human behavior capable of predicting future behavior. The most seminal work in this domain is \cite{ziebart2008maximum,ziebart2008navigate,ziebart2009human}, where Ziebart et al. propose the maximum entropy model for IRL and explore its effectiveness at predicting taxi driver routes within a city. Building on this, \cite{banovic2016modeling} leverages the maximum entropy model to identify human routines (i.e., common sequences of states and actions) from digital logs, while \cite{ziebart2012probabilistic} uses maximum entropy to predict where a person is moving their mouse cursor given a small sample of their full trajectory. Finally, the work in \cite{das2014effects} develops their own IRL model for predicting user engagement in an online community.

The second domain belongs primarily to the cognitive sciences. This work has sought to understand how individuals reason about their environments. One of the earliest results \cite{baker2007goal} showed that inferences drawn from a Bayesian-based inverse planning model (closely related to Bayesian IRL models) had a high correlation with human drawn inferences. Additional works \cite{baker2008theory-based,goodman2009cause,tauber2011using,jara-ettinger2019theory} have corroborated and extended \cite{baker2007goal} with new experiments and more sophisticated models.

The third and final domain, human-machine collaboration, has primarily been explored within the fields of machine learning and robotics where researchers have used IRL to build digital and robotic assistants that can work with humans to achieve various goals. This has led to new algorithms based on various assumption about humans agents, \cite{hadfield2016cooperative,malik2018efficient,ho2016showing}, as well as novel applications such medical robots to aid in rehabilitation \cite{woodworth2018preference}, digital assistants that provide natural language navigational directions \cite{daniele2017navigational}, and mobile robots that can navigate crowds while conforming to social expectations \cite{kretzschmar2016socially}.

The work presented here does not fit cleanly into any of the three domains as defined above. This is due to our desire to modify human behavior rather than to predict, understand, or facilitate behavior. Most notably this can be seen by the fact that we do not ever learn the reward function of our treatment targets. Rather, we only learn the reward functions for previously selected human experts. We then use the expert's reward function to nudge new participants into performing the task similar to the expert. 

Among IRL algorithms that recover what the authors consider to be full (or stand-alone) reward functions four candidates were considered, \cite{abbeel2004apprenticeship,ziebart2008maximum,levine2011nonlinear}. Of these four \cite{ziebart2008maximum} seemed most promising as it produces a single reward function and has been used extensively in human behavior studies. Even more so, \cite{levine2011nonlinear} is a kernel-based extension to \cite{ziebart2008maximum} and seemed particularly promising. Unfortunately, in experiments the authors could not get \cite{levine2011nonlinear} to complete within acceptable time limits. This was largely due to \cite{levine2011nonlinear} being linear in state, and as we will describe our state space was quite large.

Therefore, in the end, the authors used a kernel-based extension to \cite{abbeel2004apprenticeship}. It should be noted that this approach is closely related to \cite{kim2018imitation} and many of their technical arguments for their approach transfer. Our formulation differs from \cite{kim2018imitation} in that it learns the optimal policy for the current witness function on each iteration and then uses the projection method from \cite{abbeel2004apprenticeship} to calculate a new reward function. It is also worth noting that this approach is very similar to the work in \cite{jin2010gaussian}. The most notable difference is that we use the more efficient projection method from \cite{abbeel2004apprenticeship} rather than the SVM formulation.

\section{Preliminaries\label{sec:prelim}}

We assume access to a finite Markov decision process (MDP), an unknown expert reward, and an observable expert policy. That is, a tuple $(S$, $A$, $\mathcal{P}$, $d$, $R_E$, $\pi_E$, $T)$ where $S$ is a finite set of states, $A$ is a finite set of actions, $\mathcal{P}(s,a,s') = \mathrm{Pr}(s' \mid s,a)$ is a transition probability, $d(s) = \mathrm{Pr}(s)$ is an initial state distribution, $R_E: S \to \mathbb{R}$ is an unknown expert reward, $\pi_E: S \to A$ is an observable expert policy, and $T$ is the number of steps in the MDP. Additionally, the expectation with respect to an MDP for any given policy $\pi: S \to A$ is
\begin{equation}
    \mathbb{E}^{\pi}[f] = \sum_{\zeta}f(\zeta)\mathrm{Pr}(\zeta \mid \pi),
\end{equation}
where $\zeta = (s_1,s_2,\dots,s_T) \in S^T$ and $\mathrm{Pr}(\zeta \mid \pi) = d(s_1)\prod_{t=1}^{T-1}\mathcal{P}(s_t,\pi(s_t),s_{t+1})$.


\section{Algorithm \label{sec:algorithm}}

We use KPIRL to learn a reward, $R^*$, for an expert who is following an observable policy, $\pi_E$, in an MDP. KPIRL is a kernel based extension to PIRL \cite{abbeel2004apprenticeship}. In our work the expert's policy is assumed to be optimal with respect to an unknown reward $R_E$, where optimal is defined as a policy that maximizes the expected total reward,
\begin{equation}
    G(\pi, R) = \mathbb{E}^{\pi}\left[\sum_{t=1}^T R(s_t)\right]. \label{eq:optimal}
\end{equation}

To find our desired $R^*$ we begin by assuming that $R_E$ can be represented in an abstract vector space $H$ over the field $\mathbb{R}$ as an inner product of some $w \in H$ and $\phi: S \to H$ giving
\begin{equation}
    R_E(s) = \langle \phi(s), w \rangle \label{eq:rwd}.
\end{equation}
That is, $R_E \in H^*$ where $H^*$ is the dual of $H$. In this formulation we refer to $\phi(s)$ as the features of state $s$. From the linearity of the inner product the total expected reward for any policy pursuing a reward in $H^*$ can be shown to be a function of feature expectation:
\begin{equation}
    \mu_{\pi} = \mathbb{E}^{\pi}\left[\sum_{t=1}^T\phi(s_t)\right] \label{eq:feat_exp}.
\end{equation}
In other words, 
\begin{equation}
    G(\pi, s \mapsto \langle \phi(s), w \rangle) = \langle \mu_{\pi}, w \rangle.\label{eq:feat_exp_opt}
\end{equation}
In practice we estimate the feature expectation for any given policy using a sample mean. When we do this we indicate that the sample mean estimate is used via a hat such as $\hat{\mu}_\pi$.

We are now ready to introduce the kernel extension. In the original PIRL paper the vector space $H$ was assumed to be an $n$-dimensional Euclidean space, $\mathbb{R}^n$, and the inner product was the dot product. This formulation can be extended by defining $H$ to be, 
\begin{equation}
H := \mathrm{Span}\{k(s,\cdot) : s \in S\},
\end{equation}
where $k:S\times S \to \mathbb{R}$ is a positive-definite kernel (i.e., $\sum_s^S \sum_{s'}^S c_{s}c_{s'}k(s,s') \geq 0$ given $c_s,\dots,c_S \in \mathbb{R}$). We require $k$ to be positive-definite in order to define an inner product for $H$. That is, $\langle k(s_1,\cdot), k(s_2,\cdot) \rangle := k(s_1,s_2)$. This product is necessary for Equations~\ref{eq:rwd}~and~\ref{eq:feat_exp_opt}.

Using the above formulations we introduce KPIRL in Algorithm~\ref{alg:KPIRL}. KPIRL uses an iterative max-margin approach to find reward functions. On each iteration $i$ KPIRL finds a new reward $R_i$ (Line~\ref{ln:rwd}) that maximizes the margin between the expert policy, an optimal policy for the reward from the previous iteration (Line~\ref{ln:opt}), and a smoothed combination of all optimal policies up to this iteration (Line~\ref{ln:smooth}). 

KPIRL is not guaranteed to find a max-margin reward that produces expert like behavior. This is due to the fact that PIRL was designed to produce a smoothed policy identical to the expert rather than a single reward. For this reason the final line in KPIRL looks over all the rewards generated up to this point and selects the one which produced the most expert like behavior (Line~\ref{ln:final}). 

It is tempting to think that by representing $R$ as a kernel method finding a max-margin will become easier. This is, unfortunately, not the case due to the margin being linear with respect to reward values, $G(\pi_E,2R)-G(\pi,2R) = 2(G(\pi_E,R)-G(\pi,R))$, even if a chosen $k$ means that $R$ is non-linear with respect to $S$. Even if kernel methods do not create non-linear margins they can still provide two key benefits. 

First they allow designers to encode domain knowledge into the shape of the reward functions. In particular they allow for the design of a Mahalonobis distance between feature expectations. To see this, we need to define a vector $K$ containing all state kernel representations (i.e., $k(s,\cdot) \forall s \in S$) and a vector $F_\pi$ that contains the frequency of state visits for a given policy. With these, feature expectation can be written $\mu_\pi = \langle F_\pi, K \rangle$ and feature expectation distance becomes $||K^\top(F_{\pi_1}-F_{\pi_2})||$. This distance means we can ignore certain feature differences while accentuating others during max-margin iterations.

Second, and perhaps more importantly, when a kernel is a characteristic kernel the feature expectation distance between between two policies, $||K^\top(F_{\pi_1}-F_{\pi_2})||$, is only $0$ when $F_{\pi_1}-F_{\pi_2} = 0$ \cite{muandet2017kernel,kim2018imitation}. This means that when determining the best reward to use (Line~\ref{ln:final}) policies close to the expert have similar state frequencies to the expert. 

\begin{algorithm}
    \caption{Kernel Projection IRL (KPIRL)}\label{alg:KPIRL}
    \begin{algorithmic}[1]
        \State \textbf{Initialize:} set $R$ to a random reward
        \State \textbf{Initialize:} set $\pi^* \gets \argmax_\pi G(\pi,R)$
        \State \textbf{Initialize:} set $\mu_E \gets \hat{\mu}_{\pi_E}$ and $\bar{\mu} \gets \hat{\mu}_{\pi^*}$
        \item[]
        \For  {$i = 1 \to N$}
            \State $R_i \gets s \mapsto \langle \phi(s),\mu_E - \bar{\mu} \rangle$\label{ln:rwd} 
            \State $\pi^* \gets \argmax_\pi 
            G(\pi, R)$ \label{ln:opt} 
            \State $\mu_i \gets \hat{\mu}_{\pi^*}$
            \State $\alpha \gets \frac{\left\langle\mu_i - \bar{\mu},\ \mu_E - \bar{\mu}\right\rangle}{\left\langle\mu_i - \bar{\mu},\ \mu_i - \bar{\mu}\right\rangle}$
            \State $\bar{\mu} \gets \bar{\mu} + \alpha\left(\mu - \bar{\mu}\right)$\label{ln:smooth}
        \EndFor
        \item[]
        \State \textbf{Return:} $R_i$ for the $i$ which minimizes $||\mu_i-\mu_E||$\label{ln:final}
    \end{algorithmic}
\end{algorithm}

In our human subject experiments it was our experience that the kernel extension was necessary to produce reward functions that appeared reasonable in practice (i.e., during play testing). We believe this was due to a combination of the benefits mentioned previously. First, we were able to design kernel functions for our specific task that accentuated important differences leading to better max-margin rewards. And second, by using a characteristic kernel feature evaluating policy similarities using feature expectation guaranteed the learned policies matched the expert policies.

\subsection{Benchmarks \label{sec:17}}
Among existing nonlinear IRL algorithms the authors feel there are three notable standouts: \cite{levine2011nonlinear} which models the reward as a Gaussian process, \cite{ho2016generative} which models the reward as a neural network, and \cite{kim2018imitation} which also models the reward using kernels. As mentioned in related work the latter two algorithms in this list were not deemed appropriate for our use case due to only learning noisy reward functions. Therefore, we simply benchmark against GPIRL \cite{levine2011nonlinear} (along with PIRL \cite{abbeel2004apprenticeship} for reference).

For these benchmarks random grid world environments were created using the IRL Toolkit \cite{levine2011toolkit}. To make sure metrics weren't due to chance each metric was calculated over many random grid worlds with their mean performance and 95\% CI reported. Each grid world has $|S|=n^2$ for some $n \in \mathbb{N}$ with five actions (up, down, left, right and stay). All state-action transition probabilities are deterministic. 

\begin{figure}
{
    \centering
    \includegraphics[width=1\textwidth]{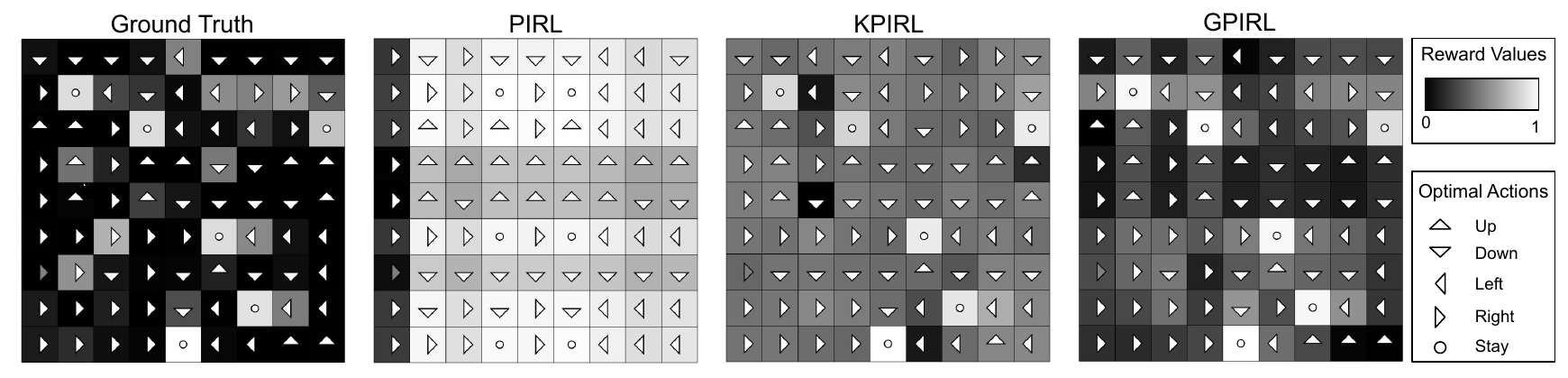}
    \caption{The $R^*$ and $\pi^*$ produced by PIRL, KPIRL and GPIRL for a ground truth. \label{fig:6}}
}
\end{figure}

To generate rewards for states a random number is drawn for each state from a uniform distribution between $0$ and $1$. This number is then raised to the power of 8 and fixed to determine each state's reward. Figure~\ref{fig:6} shows a ground truth gridworld whose reward was generated via this procedure.

To generate features for states a binary feature vector of length $2n$ is defined for each state. The first $n$ values in the vector represent rows while the final $n$ values represent columns. A state's feature vector would receive a 0 if a row or column was less than the current state's otherwise it would receive a 1. This meant the top left state always had a feature vector of all $1$'s while and the bottom right state always had a feature vector of all $0$'s except for a $1$ at array positions $n$ and $2n$. 

\begin{figure}[htb]
{
    \centering
    \includegraphics[width=1\textwidth]{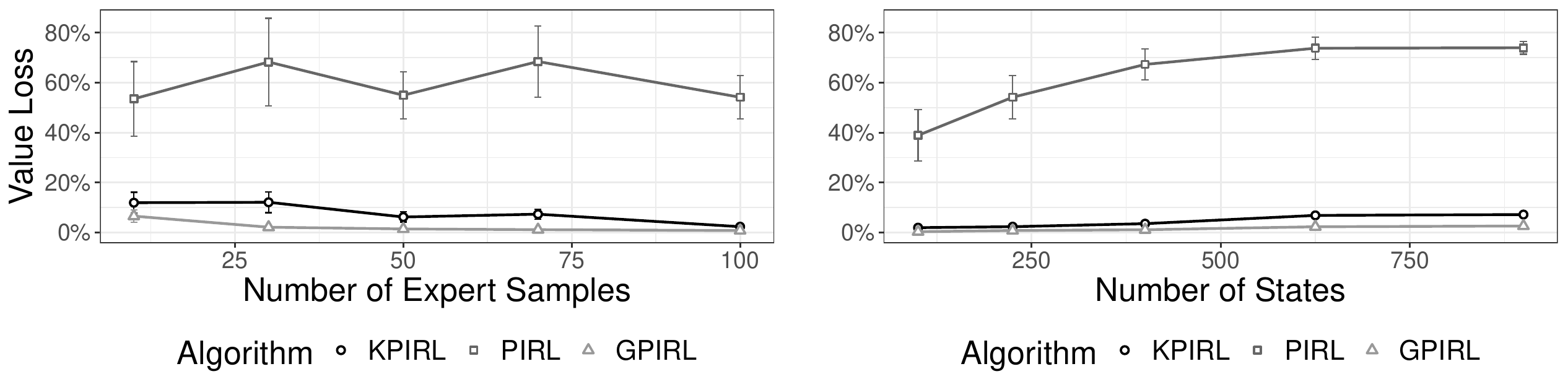}
    \caption{The \% of value lost for IRL reward policies relative to the true policy. On the left $|S|$ is fixed at 225. On the right the number of expert samples is fixed at $100$.\label{fig:5}}
}
\end{figure}

In terms of the percent of value lost, no matter the size of the state space or the number of expert trajectories, GPIRL always performed best (see Figure~\ref{fig:5}). KPIRL came in a close second, typically about 5\% behind GPIRL. The linear algorithm, PIRL, performed worst.

The experiments above were conducted primarily to show that our KPIRL algorithm was a reasonable approach in the experiment. These benchmark experiments do not show that an alternative IRL algorithm wouldn't have worked better in the human subjects experiment. Finally, while kernel methods are not strictly necessary to achieve complex function approximation, it was our experience that it was easier to find reward functions that appeared subjectively good upon inspection by starting with intuitive features and then adding complex structure through kernel methods.

\section{Experiments \label{sec:experiment}}

Two human subject experiments were conducted to test the proposed process for human apprenticeship learning via IRL. Following the proposed process (see Figure~\ref{fig:proc}) we first define a task. After that we describe how we learned reward functions from human experts. Next, we describe how reward functions were shown to the human learners. Finally, we analyze the effect of the shown rewards on the human learners.



\subsection{Task Definition \label{sec:task}}
For our human apprenticeship experiment we created a browser-based web game. The game was designed to be as simple as possible, while still validating our proposed process, so as to minimize the potential for confounding explanations of our results. 

\begin{figure}
{
    \centering
    \includegraphics[width=.95\textwidth]{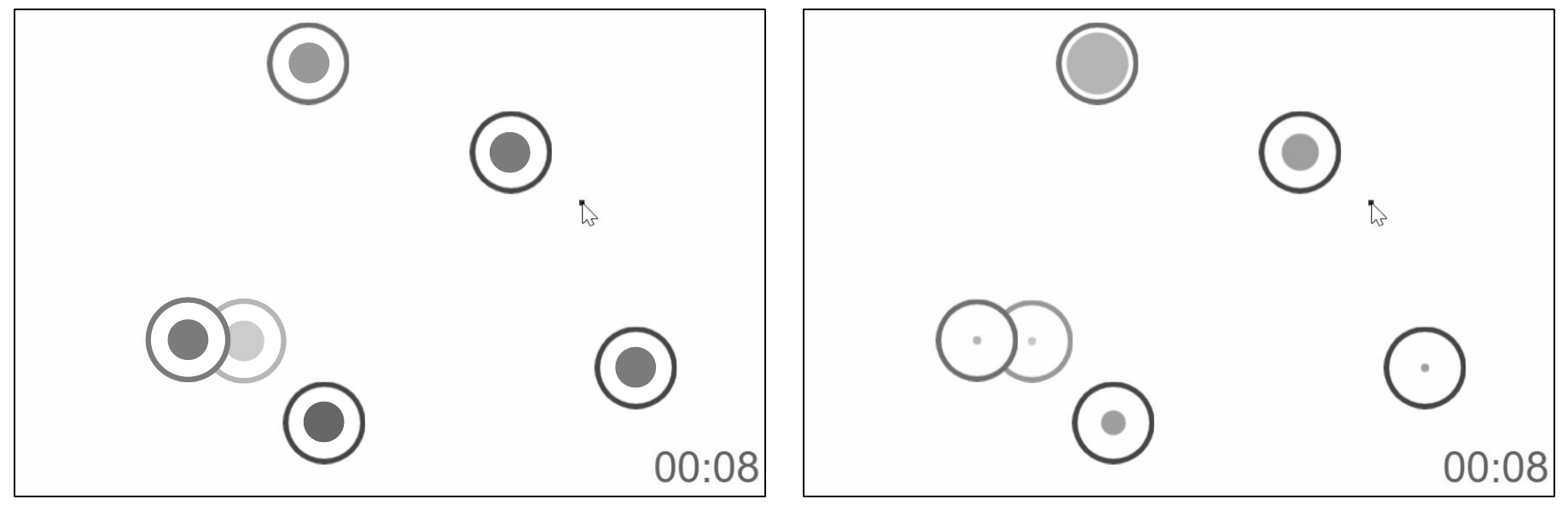}
    \caption{The left image shows a control game (i.e. a pretest game or a posttest game for the control group). The right image shows a game with a treatment reward applied. The more filled in a target the more points it was worth. All games were drawn in grayscale. \label{fig:game}}
}
\end{figure}

The game lasted for a total of $15$ seconds. During the game participants would move their mouse cursor to touch targets that randomly appeared. When a participant touched a target they would earn points equal to the target's value (see Figure~\ref{fig:game} for screenshots from two separate games).

Targets appeared following a Poisson process (with $\lambda = 5$ targets per second), were placed according to a uniform distribution (so long as they were within five pixels of screen edge), and disappeared after one second. The rate of target appearance along with their placement made it impossible for participants to touch every target.

Target point values were indicated by how filled in the inside of a target was. A fully filled in target was worth $2$ points while an empty target was worth $0$ points (once again, see Figure~\ref{fig:game} for examples of target filling). Filling in the center was used to communicate point value to avoid confounding factors such as colorblindness or screen color differences. 

Target size was always scaled so that a target's area equaled $1.57$\% of the entire playing field's area. This scaling controlled for any differences in participant device screen size or resolution. This same scaling also applied if a participant did not have their browser window maximized for some reason.

To perform well participants had to balance three competing objectives: speed, control, and planning. Regarding speed, if a participant moved too slowly they may not touch as many targets as possible. On the other hand, if they moved too quickly they could lose control of their cursor and move less efficiently. Finally, if a player didn't plan well they could touch too many low-value targets or too few high-value targets to achieve the best possible score.



\subsection{Reward Learning \label{sec:rwd_learn}} 
To learn rewards for our experiment we selected two human experts. Two were chosen, rather than one, to evaluate whether rewards learned from different experts would drive different learning outcomes. The first expert was selected due to demonstrating a high number of target touches when playing the game while the second expert was selected due to demonstrating a low number of target touches as shown in Table~\ref{tab:experts}. We will refer to these experts, respectively, as expert H and expert L.

\begin{table}
    \centering
    \caption{Descriptive statistics for the two experts in the experiment.}{
    \begin{tabular}{ c c c c l l l }
        \hline
         ID & Gender & Age & Input & $\mathbb{E}[\textrm{Touches}]$ & $\mathbb{E}[||\dot{x},\dot{y}||_2]$ & $\mathbb{E}[||\ddot{x},\ddot{y}||_2]$\\ 
         \hline
         H & Male & $25$-$34$ & Mouse & $61.05$ & $9.872$ & $5.767$ \\
         L & Male & $25$-$34$ & Mouse & $20.55$ & $2.703$ & $2.139$ \\
         \hline
    \end{tabular}}
    \label{tab:experts}
\end{table}

In addition to selecting the two experts a kernel function $k: S \times S \to \mathbb{R}$ was designed specifically for learning rewards for the web-game described in Section~\ref{sec:task}. Through experimentation we selected an RBF kernel, 
\begin{equation}
    k(s_1,s_2) = \mathrm{exp}(-\mathcal{M}(s_1,s_2)),    
\end{equation}
with $\mathcal{M}$ defined as
\begin{align}
    \mathcal{M}(s_1,s_2) = \begin{cases} 0, &\text{if $s_1$ and $s_2$ do not touch targets} \\ ||f(s_1)-f(s_2)||_2, &\text{if $s_1$ and $s_2$ do touch targets} \\ \infty, &\text{otherwise} \end{cases}\label{eq:metric}
\end{align}
and $f$ defined as 
\begin{equation}
    f(s) = \left[x, y, ||\dot{x},\dot{y}||_2, ||\ddot{x},\ddot{y}||_2, \mathrm{atan2}(\dot{x},\dot{y})\right]\label{eq:feat}.
\end{equation}

Intuitively, the kernel is best interpreted in light of the large-margin approach in KPIRL. When the feature expectation of a kernel (see Equation~\ref{eq:feat_exp}) was able to separate strategies well then the rewards we learned also better represented the expert policy we learned from.

Using the designed kernel with the two experts four reward functions were learned: HH,HL,LH, and LL. Two from expert H and two from expert L (see Table~\ref{tab:rewards}). Because KPIRL as defined in Algorithm~\ref{alg:KPIRL} produces a different reward function each time it runs we decided to learn two rewards to evaluate the stability of the treatment effect when different reward functions were learned from a single expert.

\begin{table}
    \centering
    \caption{Descriptive statistics for the rewards used in the experiment.}
    \begin{tabular}{c c c l l l}
        \hline
          Reward & Expert & $||\mu_E-\mu_{\pi^*}||_k$  & $\mathbb{E}[\textrm{Touches}]$ & $\mathbb{E}[||\dot{x},\dot{y}||_2]$ & $\mathbb{E}[||\ddot{x},\ddot{y}||_2]$\\
         \hline
         HH & H & $.1539$ & $167.113$ & $55.256$ & $92.611$\\
         HL & H & $.0705$ &   $9.774$ &  $3.592$ &  $2.422$\\  
         LH & L & $.1832$ & $162.117$ & $54.821$ & $90.171$\\  
         LL & L & $.0173$ &   $6.663$ &  $3.311$ &  $2.128$\\  
         \hline
    \end{tabular}
    \label{tab:rewards}
\end{table}

\subsection{Reward Showing \label{sec:rwd_show}}
Following our proposed process, after choosing our task and learning rewards form experts, we needed to show our learned rewards to human learners. For our task we did this by modifying the number of points targets were worth when human learners played the game.

A deeper explanation requires discussion of states, reward structure and human factors. Unfortunately, it is not immediately obvious to the authors which of these ideas may belong to a deeper theory and which are merely a consequence of our task. Therefore, we will briefly cover each significant idea for the interested reader but otherwise not dwell too long.

We have said several times now that rewards were shown through the point values of targets (e.g., see the right side of Figure~\ref{fig:game}). Unfortunately, reward is a function of states, $S$, not targets, $\mathcal{T}$. To resolve this problem we defined a function $\mathcal{N}: S \times \mathcal{T} \to S$. For our game the state produced by $\mathcal{N}$ was one where the cursor was magically placed on top of a given target without changing any movement dynamics. Using $\mathcal{N}$ the point value for a target $t \in \mathcal{T}$ in any game state $s \in S$ could be described as $R(\mathcal{N}(s,t))$.   

It is well known that given a reward function there are an infinite number of transformations that can be applied without changing the reward's optimal policy. With this in mind one might ask if some rewards within this family are more optimal for human learning than others. In our case this meant setting target values to $R(\mathcal{N}(s,t)) - R(s)$ so that targets always showed how much more reward could be earned if the player behaved differently. A consequence of this is that targets could now have negative reward values. To encourage rational behavior and remove the potential for loss aversion \cite{tversky1991loss} we simply clipped these via $\mathbf{max}\,\{R(\mathcal{N}(s,t)) - R(s),0\}$.

\begin{figure}
{
    \includegraphics[width=1\textwidth]{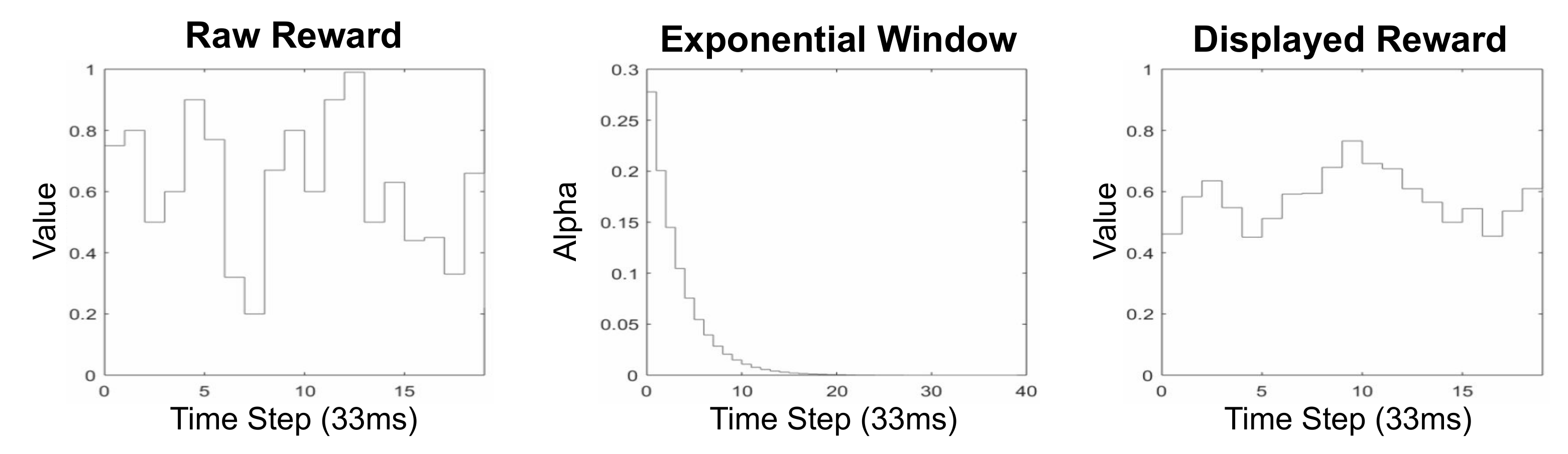}
    \caption{This shows the effect of our smoothing on the displayed reward value. On the left is the true instantaneous value $R_t$ returned by a reward function for some target. In the middle is $\alpha=5/18$ plotted as an exponential decay function. On the right is the final smoothed value, $\bar{R}_t$, which would have been displayed to participants. \label{fig:exp_smooth}}
}
\end{figure}

In a final step to make reward functions easier to interpret a dynamic transformation was also applied to rewards during the course of a game. To explain this consider some target with a displayed reward of $\bar{R}_t$ at time $t$. At the next time step this target would have a new instantaneous reward value $R_{t+1}$. To determine the next display value $\bar{R}_{t+1}$, the value $R_{t+1}$ would be smoothed with $\bar{R}_t$. This smoothing prevented sudden changes in display values. That is, $\bar{R}_{t+1} = \bar{R}_t + \alpha(R_{t+1}-\bar{R}_t)$ with $\alpha = \frac{5}{18}$. This $\alpha$ was selected via researcher play testing. An example of this smoothing is provided in Figure~\ref{fig:exp_smooth}.

\subsection{Experiment Design}
As mentioned in the introduction to Section~\ref{sec:experiment} two experiments were conducted to test our proposed process (see Figure~\ref{fig:proc}). The two experiments were identical except for the manner in which participants were assigned to treatment groups. Therefore, it is easiest to think of the second experiment as a replicate of the first, with all participants following the same protocol: (1) recruitment, (2) assignment, (3) demographics, (4) directions, and (5) observe. In what follows, details about each step in the protocol are provided where relevant.



\subsubsection{Recruitment}
Between both experiments $3,099$ participants were recruited through Amazon Mechanical Turk (AMT), an online marketplace for hiring short-term ``workers." The text used to describe the work requests was the same in both experiments and is in Table~\ref{tab:amt}. The compensation for the work requests was \$$0.20$ for experiment 1 and \$$0.10$ for experiment two. Compensation was reduced in experiment $2$ so that more participants could be recruited given a limited budget.

\begin{table}
    \centering
    \caption{AMT workers can see a work request's title, description and maximum time when choosing which requests to fulfill. These values were the same for both experiments.}{
    \begin{tabular}{ c l }
        \hline
        \textbf{Title} & \small Play two, quick (15 second) cursor (finger) touch games\\
        \textbf{Description} & \small Navigate to website to complete two 15 second cursor (finger) games.\\
        \textbf{Max Time} & $10$ Minutes\\
        \hline
    \end{tabular}}
    \label{tab:amt}
\end{table}

\subsubsection{Assignment\label{sec:assignment}}
Upon recruitment participants were assigned to one of five treatment groups. That is, the control group or one of the reward groups (see Table~\ref{tab:rewards}). A participant's assigned group determined which reward was shown to them in the posttest. In the first experiment assignment was done in batches (i.e., we assigned all participants to a group until we hit a desired quota and then began assigning to the next group). This meant that a participant's treatment group was a function of recruitment time. In the second experiment participants were assigned randomly to a treatment group at the time of recruitment. 

\subsubsection{Directions}
During the experiments participants were shown three direction screens (see Figure~\ref{fig:directions}). The first was shown at the beginning of the experiment to explain how to play the game. The second was shown before the pretest to explain that all targets would be worth one point in the next game. The third was shown before the posttest to explain that point values may vary in the next game, with the point values being indicated by how filled a target was. These three screens were always the same for all participants.

\begin{figure}
{
    \includegraphics[width=.95\textwidth]{figs/game-dir.pdf}
    \caption{These are the three direction screens shown to participants during the experiment. These  were identical for all participants, regardless of experiment and treatment.\label{fig:directions}}
}
\end{figure}

\subsubsection{Observe}
All participants were observed while playing their games. This observation took the form of recording the full game state $30$ times a second (game state included screen size, screen refresh rate, cursor location, cursor velocity, as well as the location, size, age and value of every target on the screen). Given that games were $15$ seconds long we expected to have $450$ observations per game for every participant.

\subsection{Experiment Analysis}
To test the effect of the treatments we looked at average number of targets touched. This metric was selected due to it being the differentiating factor between experts (see Table~\ref{tab:experts}). Using the metric, two hypothesis were made formed and tested in the experiments.
\begin{description}
    \item[H1] Showing a reward learned from expert H will increase the number of targets touched
    \item[H2] Showing a reward learned from expert L will decrease the number of targets touched
\end{description}

\subsubsection{Data Cleaning \label{sec:cleaning}}
At the end of the two experiments a total of $6,079$ participants had been observed. Before analysis this was filtered down to $3,060$ participants via four filters. First, if a participant enrolled more than once only the first enrollment data is kept ($n=590$) to avoid learning effects. Next, only participants who used a mouse were kept ($n=2,276$) since the reward experts used a mouse (see Table~\ref{tab:experts}). After, we removed participants for whom we did not observe all $450$ game states ($n=67$). Finally, we removed participants whose games had a frame rate below $20$ frames per second ($n=86$).

\subsubsection{Demographics Analysis}
Demographic and device information was collected for all participants. This data (see Table~\ref{tab:demographics}) did not reveal any statistically significant differences between treatment groups (see Section~\ref{sec:assignment}) for either experiment.


\begin{table}
    \centering
    \caption{All participant demographics after cleaning (see Section~\ref{sec:cleaning}).}
    \begin{tabular}{ l c c c c c c }
        \hline
                         & CT  & HH  & HL  & LH  & LL  & Total\\
        \hline
        \T Gender        &     &     &     &     &     &     \\
        \T\T Male        & $326$ & $288$ & 373 & 380 & 355 & 1722\\
        \T\T Female      & $249$ & $256$ & 261 & 272 & 296 & 1334\\
        \T\T Other       & $1$   & $0$   & 1   & 1   & 1   & 4   \\
        \T Age           &       &       &     &     &     &     \\
        \T\T 18-24       & $82$  & $87$  & 91  & 100 & 102 & 462 \\
        \T\T 25-34       & $268$ & $235$ & 295 & 328 & 284 & 1410\\
        \T\T 35-44       & $140$ & $124$ & 127 & 123 & 146 & 660 \\
        \T\T 45-54       & $45$  & $57$  & 80  & 61  & 72  & 315 \\
        \T\T 55+         & $41$  & $41$  & 42  & 41  & 48  & 213 \\
        \T Machine       &       &       &     &     &     &     \\
        \T\T Desktop     & $285$ & $278$ & 330 & 329 & 330 & 1552\\
        \T\T Laptop      & $289$ & $260$ & 305 & 317 & 322 & 1493\\
        \T\T Other       & $2$   & $6$   & 0   & 7   & 0   & 15  \\
        \T Browser       &       &       &     &     &     &     \\
        \T\T Chrome      & $480$ & $433$ & 534 & 559 & 532 & 2538\\
        \T\T Firefox     & $64$  & $77$  & 68  & 64  & 88  & 361 \\
        \T\T Other       & $32$  & $34$	 & 33  & 30  & 32  & 161 \\
        \hline
    \end{tabular}
    \label{tab:demographics}
\end{table}

\subsubsection{Pretest Analysis}
\begin{figure}
{
    \includegraphics[width=.95\textwidth]{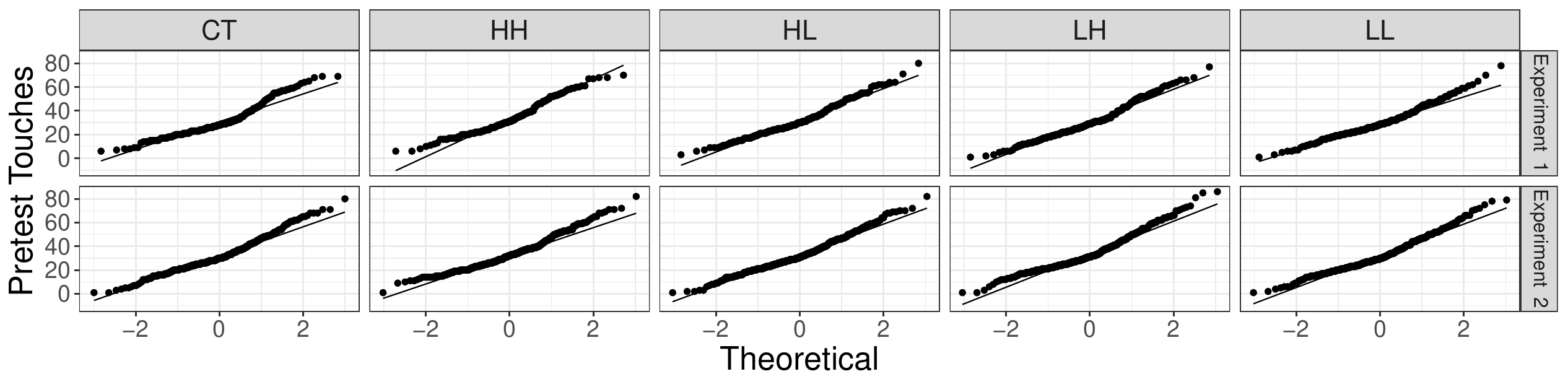}
    \caption{Experiment 1 and 2 pretest touch Q-Q plots compared to normal.\label{fig:qq}}
}
\end{figure}

To rule out whether posttest outcomes are due to treatment group differences we test for differences in pretest touches. To determine the appropriate test we check test assumptions. Given the approximate normal distribution (see Figure~\ref{fig:qq}) as well as unequal variances and sample sizes in experiment $1$ (see Table~\ref{tab:pretest}) a Welch's one-way ANOVA is selected. This gives $p=0.0676$ and $p=0.2185$ for experiment $1$ and $2$. As desired no statistically significant difference between the group means is observed.

\begin{table}
    \centering
    \caption{Summary statistics for pretest touches by treatment group.\label{tab:pretest}}.
    \begin{tabular}{ l c c c c c | c c c c c }
        \hline 
                      &  \multicolumn{5}{c}{Experiment 1} &  \multicolumn{5}{c}{Experiment 2} \\
                      & CT  & HH  & HL  & LH  & LL & CT  & HH  & HL  & LH  & LL\\
        \hline
        Mean          & 31.3 & 34.3 & 32.3 & 30.9 & 30.2 & 32.4 & 33.7 & 33.1 & 34.4 & 32.6 \\
        SD            & 13.4 & 14.6 & 13.6 & 14.3 & 12.6 & 13.6 & 13.1 & 13.5 & 14.2 & 13.6 \\
        N             & 213 & 150 & 219 & 230 & 257 & 363 & 394 & 416 & 423 & 395 \\

        \hline
    \end{tabular}
\end{table}

\subsection{Posttest Analysis}
Having ruled out baseline group differences in the pretest analysis we are ready to look at posttest outcomes in order to test hypotheses H1 and H2.

\begin{figure}
{
    \centering
    \includegraphics[width=1\textwidth]{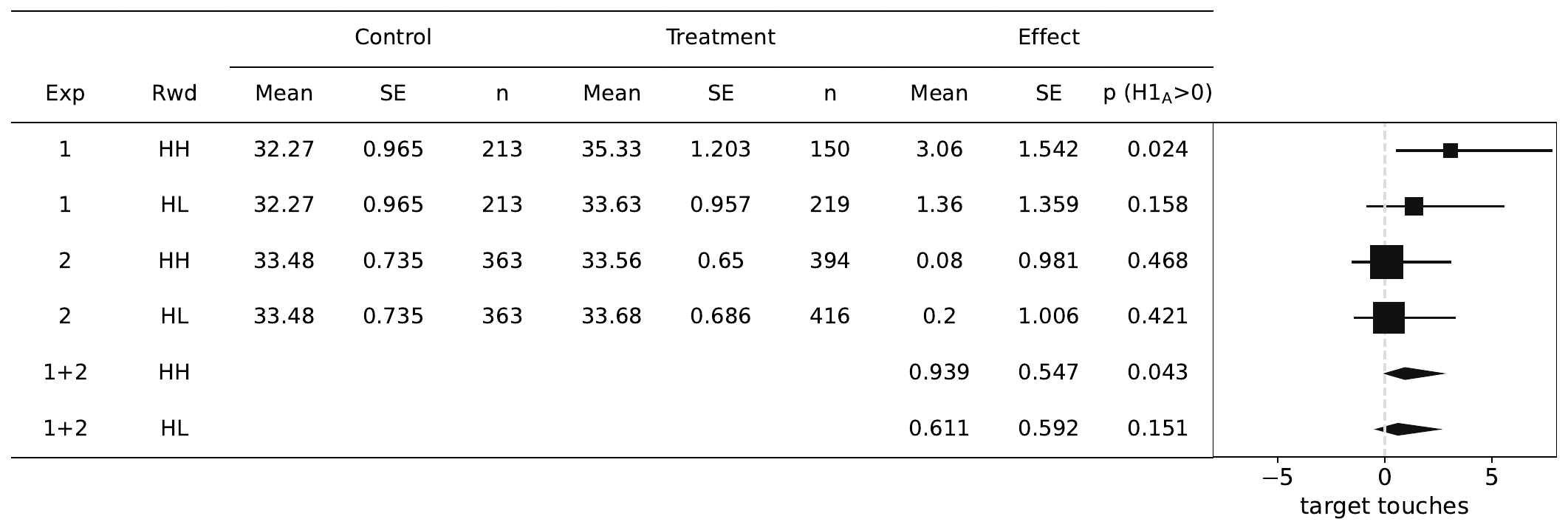}
    \caption{Forest Plot of reward HH and HL effects in both experiments. \label{fig:forest_H1}}
}
\end{figure}

For H1 we observed consistent, supporting evidence in both experiment $1$ and $2$ (see Figure~\ref{fig:forest_H1}). In experiment 1 we observed an effect of $3.06$ ($p=0.024$, $\text{SE}=1.542$) for reward HH and $1.36$ ($p=0.158$, $\text{SE}=1.359$) for reward HL. In experiment $2$ we observed a weaker but still positive effect of $.08$ ($p=0.468$, $\text{SE}=.981$) and $0.2$ ($p=0.421$, $\text{SE}=1.006$) respectively. Combining these effects via a fixed-effects model\cite{borenstein2010basic} gives overall estimates of $0.939$ ($p=0.043$, $\text{SE}=0.547$) and $9.611$ ($p=0.151$, $\text{SE}=0.592$) respectively. Given the consistency of the evidence we reject the null and accept hypothesis H1.

\begin{figure}
{
    \centering
    \includegraphics[width=1\textwidth]{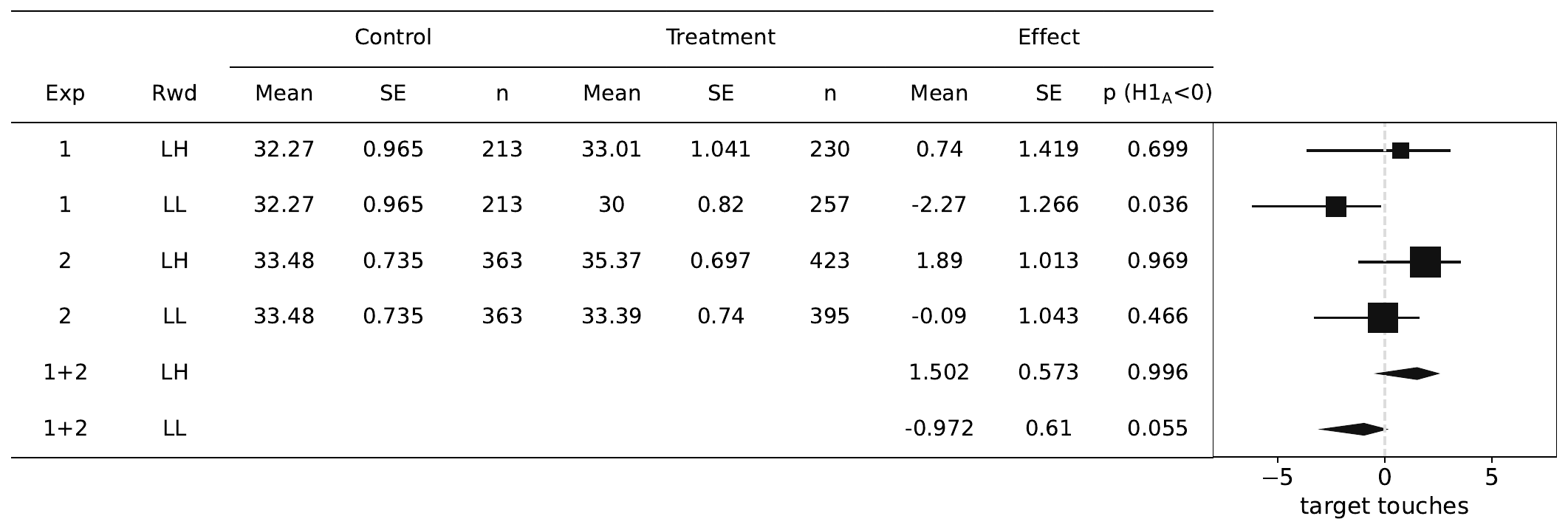}
    \caption{Forest Plot of reward LH and LL effects in both experiments.\label{fig:forest_H2}}
}
\end{figure}

Moving on to hypothesis H2 the evidence was mixed in both experiments (see Figure~\ref{fig:forest_H2}). For experiment $1$ reward LH had an effect opposite of the hypothesized direction with a mean effect of $0.74$ ($p=0.699$, $\text{SE}=1.419$) while reward LL was in the hypothesized direction with $-2.27$ ($p=0.036$, $\text{SE}=1.266$). In experiment $2$ the direction of effect remained the same for both rewards with $1.89$ ($p=0.969$, $\text{SE}=1.013$) and $-0.09$ ($p=0.466$, $\text{SE}=1.043$) respectively. Combining these via a fixed-effects model our best estimate of the reward effects is $1.502$ ($p=0.996$, $\text{SE}=0.573$) and $-0.972$ ($p=0.055$, $\text{SE}=0.61$). Given this mixed evidence we fail to reject the null for hypothesis H2.

The observed positive effect for reward LH is interesting. The authors believe this occurred due to LH representing an ``easier" policy (i.e., learned from a slower expert, see Table~\ref{tab:experts}) while still achieving many touches (see Table~\ref{tab:rewards}). However, that theory is counteracted by the fact that LH and HH have nearly equal average velocity and acceleration for their optimal policies (see Table~\ref{tab:rewards}). This suggests that there is more going on than this study could capture.

\section{Conclusion\label{sec:conclusion}}

This paper started with a simple question: could an IRL reward function learned from human experts be used to teach other players more quickly than players could learn on their own. We called such an approach human apprenticeship learning due to its relationship to digital agents that learn from human experts. By conducting two human subjects experiments we were able to show with statistical significance that the answer is yes. As often happens during the course of the research many more unanswered questions were revealed.

Perhaps the most notable open question that we discovered was what constraints and assumptions should be used when developing IRL algorithms for this use case. Recent work in apprenticeship learning (e.g., \cite{ho2016generative,kim2018imitation}) has sought to develop algorithms which directly learn policies that match some measure of feature expectation. For human apprenticeship learning these algorithms seem to be  unusable as we need the reward function to use for human feedback rather than a policy. Therefore, more efficient algorithms that directly produce reward functions would be of great use.

A second algorithmic problem that came up was determining what reward functions were appropriate for human learning. We described above several post-processing steps we followed to prepare our reward function for human consumption. It would be valuable to  be able to define algorithmic constraints (perhaps through some kind of a regularization function) that directly learn usable reward functions for this purpose. Beyond the algorithmic considerations though it is not immediately obvious from this work what those constraints should look like.

In addition to the algorithmic questions it was also clear from the experiments that many questions about human factors remain unanswered. This can be seen in the variance of observed effect sizes for the given reward functions. While the effect of rewards on performance preserved their directions between experiments, the magnitude of the effects changed considerably. The experimental design here was not appropriately constructed to even offer suggestions for why this was. Therefore, there seems to be considerable room for improvements in future work for describing and modeling the human factors that may correlate with reward effects.

Finally, an important question in research regarding human learning or human behavior changing is long-term effects. These experiments demonstrated that in a very small time window the reward based interventions had the hypothesized effects. Repeating this study over a longer time period would be a valuable experiment to see if humans can internalize these reward functions in meaningful and lasting sense. 

In conclusion, the effect sizes of the experiments were small but significant. The results were seen as even more of a success given that experiments were not conducted in a controlled lab setting, but rather on remote devices in uncontrolled environments by unknown participants. There remains room for improvement, but all these improvements appear to be within the reach of current technology and experimental methods. Most notably future algorithmic advancements which can reduce researcher intervention in the design of reward feedback, future modelling and human factors work to better explain the experimental results and new experiments testing long-term effects of this novel intervention all seem like valuable avenues of research.

\vskip 0.2in
\bibliography{_main.bib}

\end{document}